\documentclass[lettersize,journal]{IEEEtran}

\usepackage{amsmath,amssymb}
\usepackage{graphicx}
\usepackage{url}
\usepackage{booktabs}
\usepackage{makecell}
\usepackage{colortbl}
\usepackage{stfloats}
\usepackage{balance}

\usepackage{algorithmic}

\usepackage[caption=false,font=normalsize,labelfont=sf,textfont=sf]{subfig}

\usepackage[colorlinks=true, linkcolor=blue, filecolor=magenta, citecolor=blue, urlcolor=black]{hyperref}
\usepackage[capitalise]{cleveref}

\usepackage{orcidlink}

\usepackage{bm}

\begin{document}

\title{OccludeNet: A Causal Journey into Mixed-View Actor-Centric Action Recognition under Occlusions}

\author{%
 Guanyu~Zhou\orcidlink{0009-0003-9781-6646}, 
 Wenxuan~Liu\orcidlink{0000-0002-4417-6628},~\IEEEmembership{Member,~IEEE},
 Wenxin~Huang\orcidlink{0000-0001-8683-7327},~\IEEEmembership{Member,~IEEE}, 
 Xuemei~Jia\orcidlink{0000-0003-1762-4365},
 Xian~Zhong\orcidlink{0000-0002-5242-0467},~\IEEEmembership{Senior~Member,~IEEE},
 Chia-Wen~Lin\orcidlink{0000-0002-9097-2318},~\IEEEmembership{Fellow,~IEEE}
 \thanks{Manuscript received June 7, 2025. This work was supported in part by the National Natural Science Foundation of China under Grants 62271361 and 62301213, and in part by the Hubei Provincial Key Research and Development Program under Grant 2024BAB039. (Corresponding author: Xian Zhong.)}
 

 \thanks{Guanyu Zhou and Xian Zhong are with the Sanya Science and Education Innovation Park, Wuhan University of Technology, Sanya 572025, China, and the Hubei Key Laboratory of Transportation Internet of Things, Wuhan University of Technology, Wuhan 430070, China (e-mail: guanyuzhou.ai@gmail.com; zhongx@whut.edu.cn).}
 
 \thanks{Wenxuan Liu is with the State Key Laboratory for Multimedia Info Processing, Peking University, Beijing 100871, China (e-mail: lwxfight@126.com).}
 
 \thanks{Wenxin Huang is with the Hubei Key Laboratory of Big Data Intelligent Analysis, Hubei University, Wuhan 430062, China (e-mail: wenxinhuang\_wh@163.com).}
 
 \thanks{Xuemei Jia is with the National Engineering Research Center for Multimedia Software, Wuhan University, Wuhan 430072, China (e-mail: jiaxuemeil@whu.edu.cn).}
 
 
 \thanks{Chia-Wen Lin is with the Department of Electrical Engineering, National Tsing Hua University, Hsinchu 30013, Taiwan (e-mail: cwlin@ee.nthu.edu.tw).}
}

\maketitle

\begin{abstract}
The lack of occlusion data in common action recognition video datasets limits model robustness and hinders consistent performance gains. We build \textsc{OccludeNet}, a large-scale occluded video dataset including both real and synthetic occlusion scenes in different natural settings. \textsc{OccludeNet} includes dynamic occlusion, static occlusion, and multi-view interactive occlusion, addressing gaps in current datasets. Our analysis shows occlusion affects action classes differently: actions with low scene relevance and partial body visibility see larger drops in accuracy. To overcome the limits of existing occlusion-aware methods, we propose a structural causal model for occluded scenes and introduce the Causal Action Recognition (CAR) method, which uses backdoor adjustment and counterfactual reasoning. This approach strengthens key actor information and improves model robustness to occlusion. We hope the challenges of \textsc{OccludeNet} will encourage more study of causal links in occluded scenes and lead to a fresh look at class relations, ultimately leading to lasting performance improvements. Our code and data is availibale at: \url{https://github.com/The-Martyr/OccludeNet-Dataset}.

\end{abstract}


\section{Introduction}
\label{sec:intro}

Action recognition is essential for understanding human behavior~\cite{ivc/HerathHP17, ijcv/KongF22} and has achieved strong results on closed-set benchmarks such as \textsc{Kinetics}~\cite{corr/KayCSZHVVGBNSZ17}, \textsc{UCF101}~\cite{corr/abs-1212-0402}, and \textsc{HMDB51}~\cite{iccv/KuehneJGPS11}. However, these datasets textitasize ideal conditions by filtering out ambiguous samples, which limits their applicability in real-world scenarios where actors are often occluded.


Current occlusion video datasets lack multi-view perspectives and effective motion information. To improve the robust ability of action recognition under occlusion environment, a preliminary exploration has been conducted of \textsc{K-400-O}~\cite{nips/GroverVR23}. However, it applies large-scale occlusions to the entire video frame under a single view. This strategy of adding occlusion akin to blind video reconstruction, neglecting the interrelations among scene elements and actions. \textit{Diverse contextual elements} and \textit{class correlations} under occlusions have not received the necessary attention.

\begin{figure}[t]
	\centering
	\includegraphics[width = \linewidth]{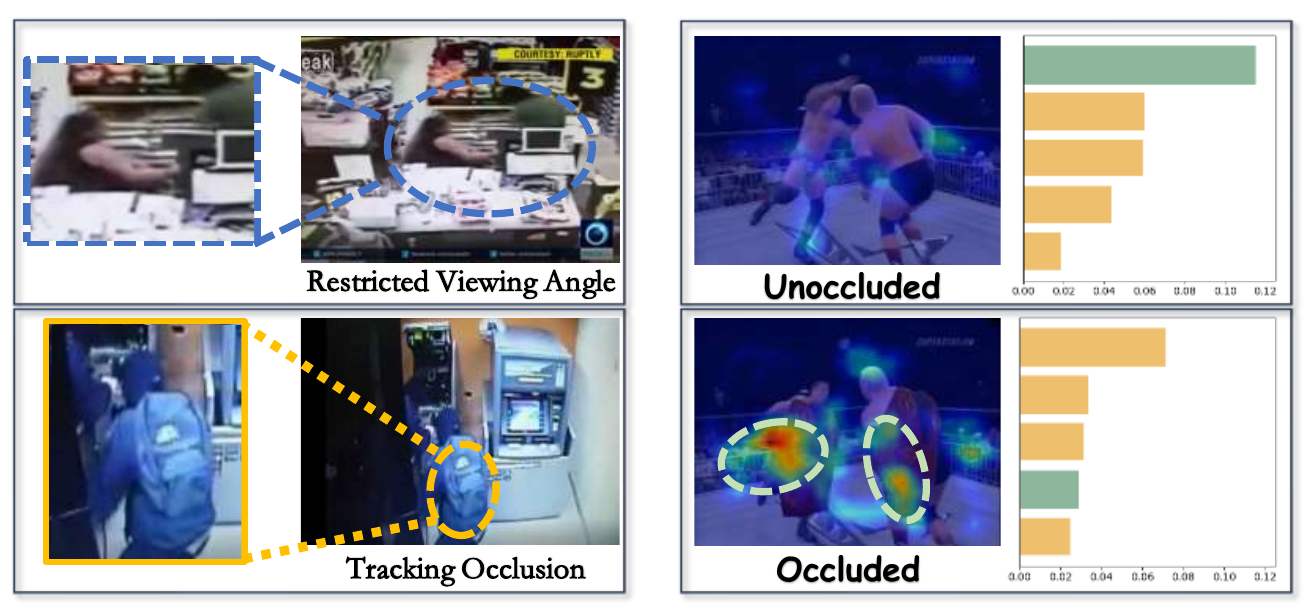}
	\caption{\textbf{Common real-world occlusions}. Left: scenes exhibiting ambiguous occlusions. Right: Grad-CAM~\cite{Selvaraju_2019} visualizations of original and occluded \textsc{Kinetics-400}~\cite{corr/KayCSZHVVGBNSZ17} samples, illustrating how occlusions misdirect model attention and degrade performance. Green columns indicate the correct class.}
	\label{fig:motivation}
\end{figure}

\begin{figure*}[t]
	\centering
	\includegraphics[width = \textwidth]{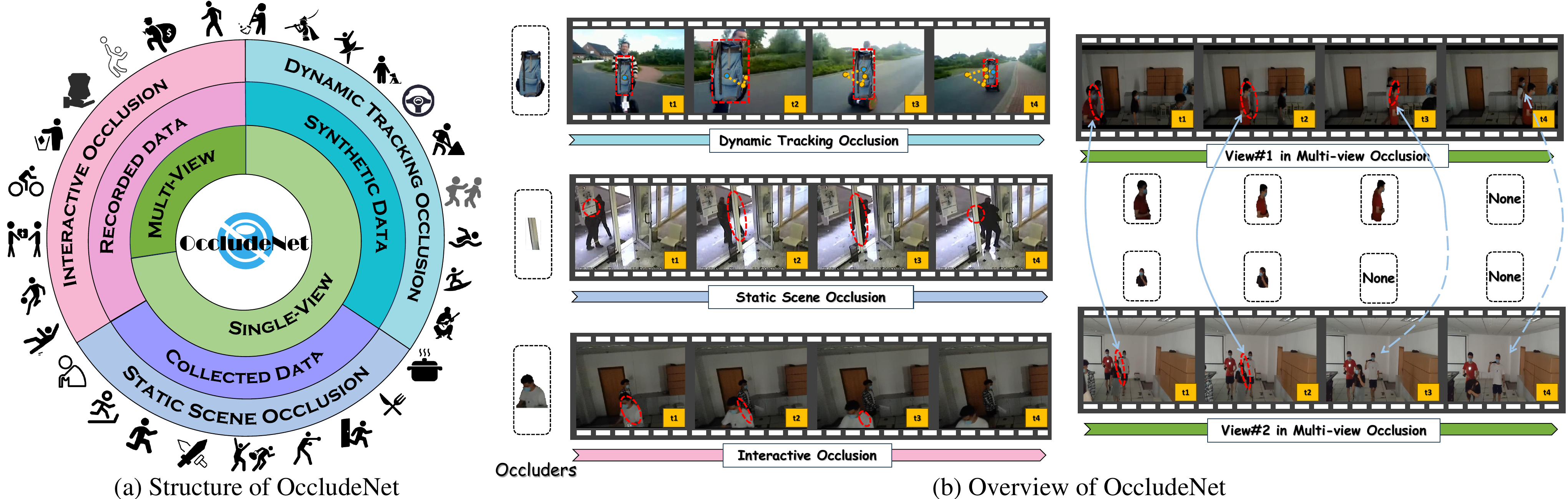}
	\caption{\textbf{\textsc{OccludeNet} dataset}. (a) {Hierarchical annotations}: Coarse-to-fine labels of contextual elements from single- and multi-view videos covering three occlusion types, dynamic tracking, static scene, and interactive. Data are drawn from real recordings, curated collections, and synthetic sources. (b) {Occlusion overview}: Dynamic tracking occlusions feature an occluder following the actor (\textcolor{blue}{blue dot} and \textcolor{orange}{orange trajectory}); static scene occlusions use stationary occluders to partially block the actor; interactive occlusions arise from actor interactions (\textcolor{red}{red dotted lines}).}
	\label{fig:OccluDet}
\end{figure*}

By analyzing existing video occlusions (see \cref{fig:motivation}), we present \textsc{OccludeNet}, a comprehensive occlusion video dataset that spans 424 action classes and incorporates diverse natural environment and occlusion types (see \cref{fig:OccluDet}). 
It comprises dynamic tracking occlusions, static scene occlusions, and interactive occlusions:
\textit{1) \textsc{OccludeNet-D}:} dynamic tracking occlusions involving moving actors;
\textit{2) \textsc{OccludeNet-S}:} static scene occlusions by background elements;
\textit{3) \textsc{OccludeNet-I}:} single-view interactive occlusions under changing lighting;
\textit{4) \textsc{OccludeNet-M}:} multi-view interactive occlusions.
By integrating both synthetic and real-world data, we enhance the dataset's validity and scalability.

Existing de-occlusion models typically focus on minimizing the impact of occluded regions. However, the boundaries between occluders, backgrounds, and actors are often ambiguous. Traditional methods tend to rely on statistical correlations~\cite{corr/abs-2307-13992}, thereby failing to capture the causal relations between occluders, backgrounds, visible actor parts, and predictions~\cite{li2023mitigating, cvpr/DingLYQXCWX22}.

We hypothesize that occlusion features serve as a confounder in the causal relation between actor attributes and model predictions~\cite{pearl2009causality}. To address this, we introduce a Causal Action Recognition (CAR) using a structural causal model that incorporates occlusion elements. Through counterfactual reasoning, our approach redirects the model's focus to the causal influence of unoccluded actor features. We quantify this causal effect by calculating the difference between predictions under occlusion and counterfactual predictions, interpreting the improvement in correct class probability as a positive treatment effect~\cite{pearl2009causality}. This method counteracts the accuracy loss due to occlusion by introducing an additional supervised signal that combines relative entropy loss with cross-entropy, thereby enhancing the model's robustness to occlusions across various datasets.

In summary, our contributions are threefold:

\begin{itemize}
	\item We construct an innovative dataset \textsc{OccludeNet}, designed specifically for occlusion-aware action recognition. It \textbf{uniquely} covers diverse occlusion scenarios, including \textit{dynamic tracking} occlusions, \textit{static} scene occlusions, and \textit{multi-view interactive} occlusions. \textsc{OccludeNet} derives its strength from its comprehensive coverage of occlusion types, making it highly effective in addressing real-world problems with significant research value and practical applicability.
 
	\item We propose an insightful causal action recognition (CAR) framework that directs models to focus on unoccluded actor parts, enhancing robustness against occlusions.
 
	\item Our study reveals that occlusion strategies centered on actors influence action class recognition to varying degrees, highlighting the necessity of tailored approaches for addressing occlusions.

\end{itemize}

\begin{table*}[t]
	\centering
	\setlength{\tabcolsep}{4pt}
	\caption{\textbf{Comparison of \textsc{OccludeNet} with existing occluded video action recognition datasets}. Syn.: synthesized; O-D: occlusion duration ratio; ``Uncropped'' indicates availability of full-frame clips; ``Inter-class'' denotes class correlation profiling.}
	\begin{tabular}{l|ccccccccccc}
	\toprule[1.1pt]
	Dataset & Type & \#Clips & \#Classes & \#Occluders & O-D & View & Dynamics & Duration (s) & FPS & Uncropped & Inter-class \\
	\midrule
	\textsc{K-400-O}~\cite{nips/GroverVR23}	& Syn. & 40,000	& 400 & 50	& 100\%	& Single-view & Geometric	& 12.0	& 24	& $\times$ & $\times$ \\
	\textsc{UCF-101-O}~\cite{nips/GroverVR23}	& Syn. & 3,783	& 101 & 50	& 100\%	& Single-view & Geometric	& -	& raw	& $\times$ & $\times$ \\
	\textsc{UCF-19-Y-OCC}~\cite{nips/GroverVR23} & Real & 570	& 19	& -	& -	& Single-view & -	& 4.0-5.0	& 23.98-30& $\times$ & $\times$ \\
	\rowcolor{gray!20}
	\textsc{OccludeNet-D}	& Syn. & 233,769 & 400 & 2,788& 0-100\%	& Single-view & Tracking	& 0.50-10.15 & 6-30	& $\checkmark$ & $\checkmark$ \\
	\rowcolor{gray!20}
	\textsc{OccludeNet-S}	& Real & 256	& 5	& Varied& 0-100\%	& Single-view & Static	& 8.0-10.0	& 25-30	& $\checkmark$ & $\checkmark$ \\
	\rowcolor{gray!20}
	\textsc{OccludeNet-I}	& Real & 345	& 7	& Varied& 0-100\%	& Single-view & Interactive & 5.0-10.0	& 30	& $\checkmark$ & $\checkmark$ \\
	\rowcolor{gray!20}
	\textsc{OccludeNet-M}	& Real & 1,242	& 12	& Varied& 0-100\%	& Multi-view	& Interactive & 5.0-10.0	& 30	& $\checkmark$ & $\checkmark$ \\
	\bottomrule[1.1pt]
	\end{tabular}
	\label{table1}
\end{table*}

\section{Related Work}

\subsection{Action Recognition under Occlusion}

Recent methods for action recognition under occlusion aim to improve model robustness. Early work trains classifiers on independently processed HOG blocks from multiple viewpoints~\cite{eccv/WeinlandOF10} or handles blurred perspectives in multi-view settings~\cite{tip/LiuZZJWL23}. Shi \textit{et al.}~\cite{tii/ShiLWY23} introduce a multi-stream graph convolutional network to manage diverse occlusion types. Chen \textit{et al.}~\cite{chen2023unveiling} show that pre-training on occluded skeleton sequences followed by k-means clustering enhances self-supervised skeleton-based action recognition. Liu \textit{et al.}~\cite{LIU2025110948} propose a pseudo-occlusion strategy for real-world scenarios, and \textsc{Occlusion Skeleton}~\cite{bigdataconf/WuQWF20} offers a valuable resource. Benchmark studies evaluate recognizer performance under occlusion~\cite{nips/GroverVR23}. Despite these advances, annotations remain labor-intensive and explicit causal modeling of occlusion effects is absent. There remains an urgent need for large-scale occluded datasets. \cref{table1} compares \textsc{OccludeNet} with existing occlusion datasets in terms of scale, occlusion dynamics, and sample characteristics.

\subsection{Causal Inference in Computer Vision}

Causal inference has gained traction in computer vision, with applications in autonomous driving~\cite{yang2018intelligent} and robotics~\cite{pearl2009causality, corr/abs-2307-13992,xie2020robot}. It has been integrated into explainable AI~\cite{xu2021causality}, fairness~\cite{xu2019achieving, zhang2018fairness}, and reinforcement learning~\cite{iclr/ZhuNC20, madumal2020explainable}. In compositional action recognition, counterfactual debiasing helps mitigate dataset biases~\cite{mm/SunWLLDG21}, and in visual question answering, it reduces spurious correlations to yield more unbiased predictions~\cite{cvpr/NiuTZL0W21, zhou2024mitigating}. In this work, we leverage causal inference to enhance robustness in action recognition under occlusions, charting a new direction for vision research.

\section{\textsc{OccludeNet} Dataset}

We introduce \textsc{OccludeNet}, a \textbf{mixed-view} (multi-view and single-view) video dataset for action recognition, which includes three common types of occlusion (see \cref{fig:OccluDet}(a)). This section provides an overview of the dataset's construction, statistics, and features.

\subsection{Dataset Construction}
\label{sec:3.1}

We construct a large-scale dataset of mixed-view occlusion videos captured under diverse natural lighting conditions, combining both synthetic and real-world data for enhanced validity and scalability. The dataset centers on key motion information and comprises four subsets (see \cref{fig:OccluDet}(b)):

\begin{figure}[t]
	\centering
	\includegraphics[width = \linewidth]{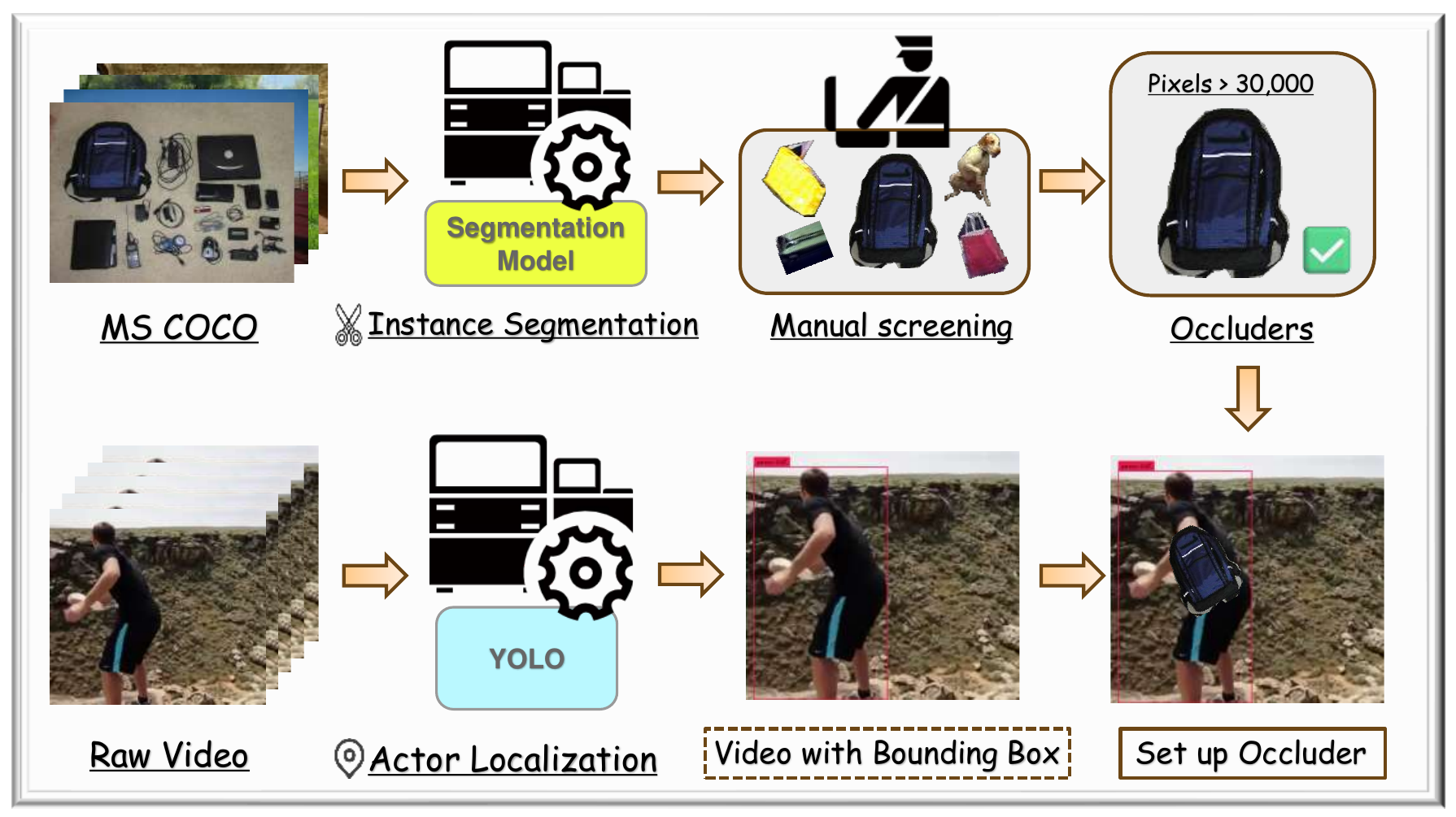}
	\caption{\textbf{Construction process of \textsc{OccludeNet-D}}. For each frame, we detect the actor's bounding box using YOLOv5~\cite{glenn_jocher_2020_4154370}, center a random occluder on that box, and dynamically scale it to simulate tracking occlusion.}
	\label{fig:process}
\end{figure}


\subsubsection{\textsc{OccludeNet-D} (Dynamic Tracking Occlusion)}	

For dynamic tracking occlusion, we sample videos from \textsc{Kinetics-400}~\cite{corr/KayCSZHVVGBNSZ17} and introduce synthetic occluders as primary elements over actor bounding boxes to challenge recognition while preserving action identity. The construction process is illustrated in \cref{fig:process}. We further evaluate the generalization capacity of synthetic data in \ref{sec:syn}.

\subsubsection{\textsc{OccludeNet-S} (Static Scene Occlusion)}	

To focus on static occlusion, we select videos from \textsc{UCF-Crime}~\cite{cvpr/SultaniCS18} that prominently feature actors partially blocked by scene elements. These are then segmented into clips matching \textsc{Kinetics-400} durations, providing realistic static-occlusion scenarios.

\subsubsection{\textsc{OccludeNet-I} (Single-View Interactive Occlusion)}	

We record continuous action sequences at $1920\times1080$ with a fixed RGB camera under varying lighting conditions. Participants with diverse body types and clothing styles wear masks for anonymity. Audio is retained for potential multi-modal analysis.

\subsubsection{\textsc{OccludeNet-M} (Multi-View Interactive Occlusion)}	

For multi-view data, we employ synchronized RGB and near-infrared cameras at three angles spaced $120^\circ$ apart to record the same action sequence. The setup and conditions mirror those of \textsc{OccludeNet-I}, ensuring uniform resolution, frame rate, and audio inclusion.

\subsection{Quality Control}

All videos undergo careful manual screening to ensure each clip accurately represents its action class with sufficient occlusion; approximately 4,000 samples were discarded due to blur or unrecognizability. We ensure diversity of actor appearance (body shape, clothing) and maintain occluder resolution above 30,000 pixels for visual clarity. All contributors (the authors) are professionally accountable for video quality. The entire dataset is uniformly annotated to facilitate efficient benchmarking across models.

\begin{figure}[t]
	\centering
	\includegraphics[width = \linewidth]{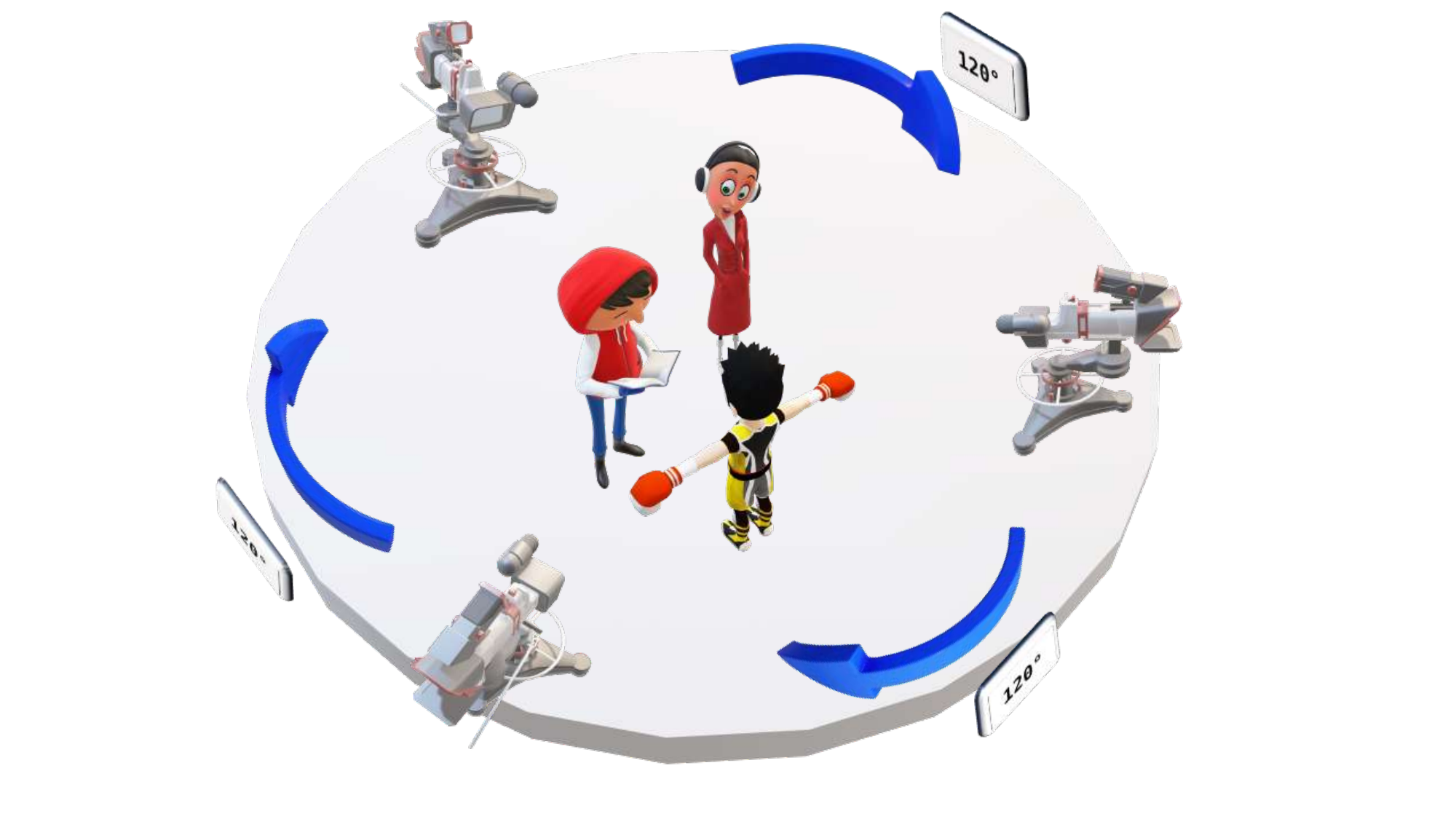}
	\caption{\textbf{Recording setup of \textsc{OccludeNet-M}}. Three cameras, positioned $120^\circ$ apart (blue arrows), simultaneously capture each action sequence from different directions.}
	\label{fig:m}
\end{figure}

\subsection{Construction Details of \textsc{OccludeNet-M}}

\cref{fig:m} presents the multi-view interactive occlusion setup. We record each action with three synchronized RGB and near-infrared camera pairs, positioned $120^\circ$ apart around the actor. Occlusion conditions vary from single-view blockages to simultaneous multi-view occlusions, spanning different environments (indoor/outdoor), motion amplitudes (small gestures to large movements), and scales (near/far). Clips range from 5 s to over 10 s at $1920\times1080$ resolution, with audio retained for multimodal analysis.

\begin{figure}[t]
	\centering
	\includegraphics[width = \linewidth]{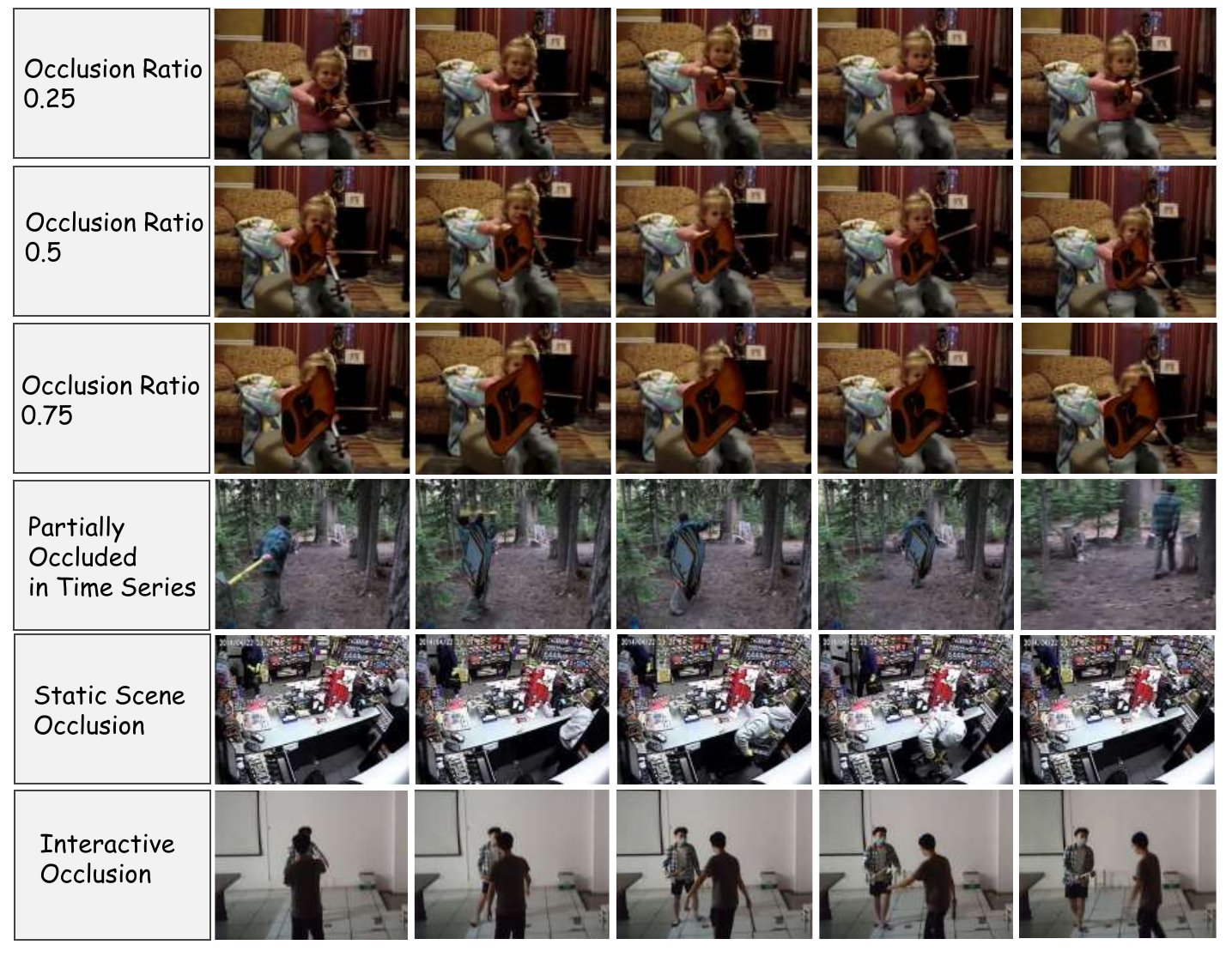}
	\caption{\textbf{\textsc{OccludeNet} samples under various occlusion conditions}. Rows 1-3 show the same original video at low, medium, and high occlusion levels using the same occluder. Row 4 illustrates progressive partial occlusion over time. Row 5 depicts actors partially blocked by static scene elements, and Row 6 shows interactive occlusions between actors.}
	\label{fig:samples}
\end{figure}

\subsection{Data Samples}

\subsubsection{Video Samples}

\cref{fig:samples} showcases example clips from \textsc{OccludeNet}, illustrating:
\textit{a) Occlusion Degree:} Light to heavy occlusion ratios.
\textit{b) Duration Ratio:} Fraction of the clip during which the actor is occluded.
\textit{c) Dynamics:} Static \textit{vs.} moving occluders; single-view \textit{vs.} multi-view interactions.

\begin{figure}[t]
	\centering
	\includegraphics[width = \linewidth]{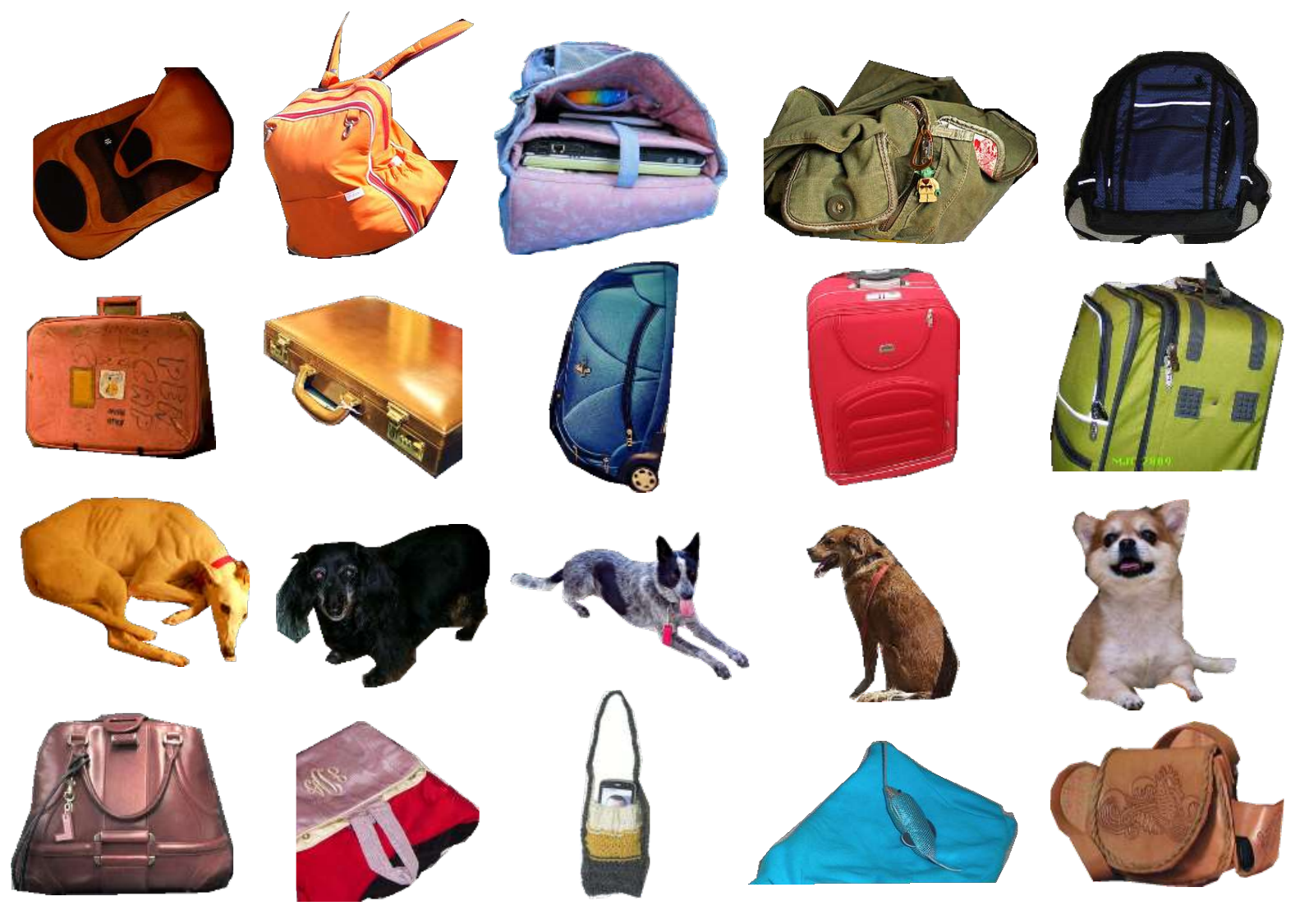}
	\caption{\textbf{Occluder examples}. Rows 1-4 illustrate the four occluder types in \textsc{OccludeNet}: backpack, dog, suitcase, and handbag, ensuring a diverse range of occlusion patterns.}
	\label{fig:occ}
\end{figure}

\subsubsection{Occluder Samples}

We select four common COCO objects~\cite{eccv/LinMBHPRDZ14}, backpacks, handbags, suitcases, and dogs, to generate dynamic occlusions. Using a segmentation model, we extract their foreground masks. Each occluder retains its original filename and pixel-level annotation for traceability (see \cref{fig:occ}).



\begin{figure}[t]
	\centering
	\includegraphics[width = \linewidth]{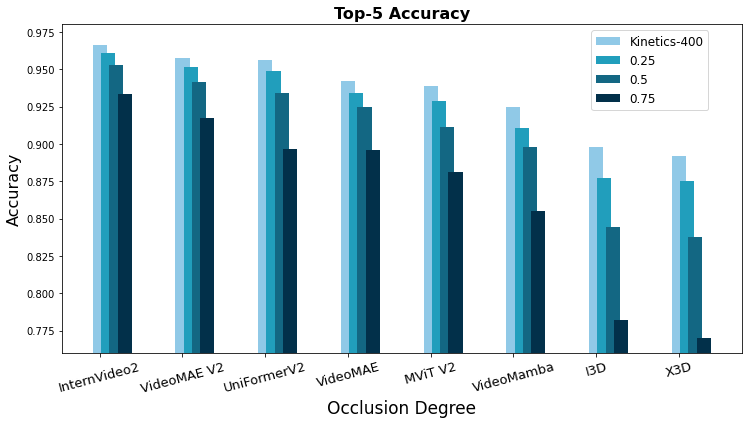} 
	\caption{\textbf{Top-5 accuracy drop on \textsc{OccludeNet} at varying occlusion degrees}. Results are shown for occlusion levels of 0.25, 0.50, and 0.75.}
	\label{fig:top5}
\end{figure}

\subsection{Additional Results and Visualizations}

\subsubsection{Benchmarking and Analysis}

\cref{fig:top5} plots Top-5 accuracy for seven models on \textsc{OccludeNet}. All models' performance degrades as occlusion increases; CNN-based architectures suffer the largest drops, while transformer-based models show greater resilience.

\begin{figure}[t]
	\centering
	\includegraphics[width = \linewidth]{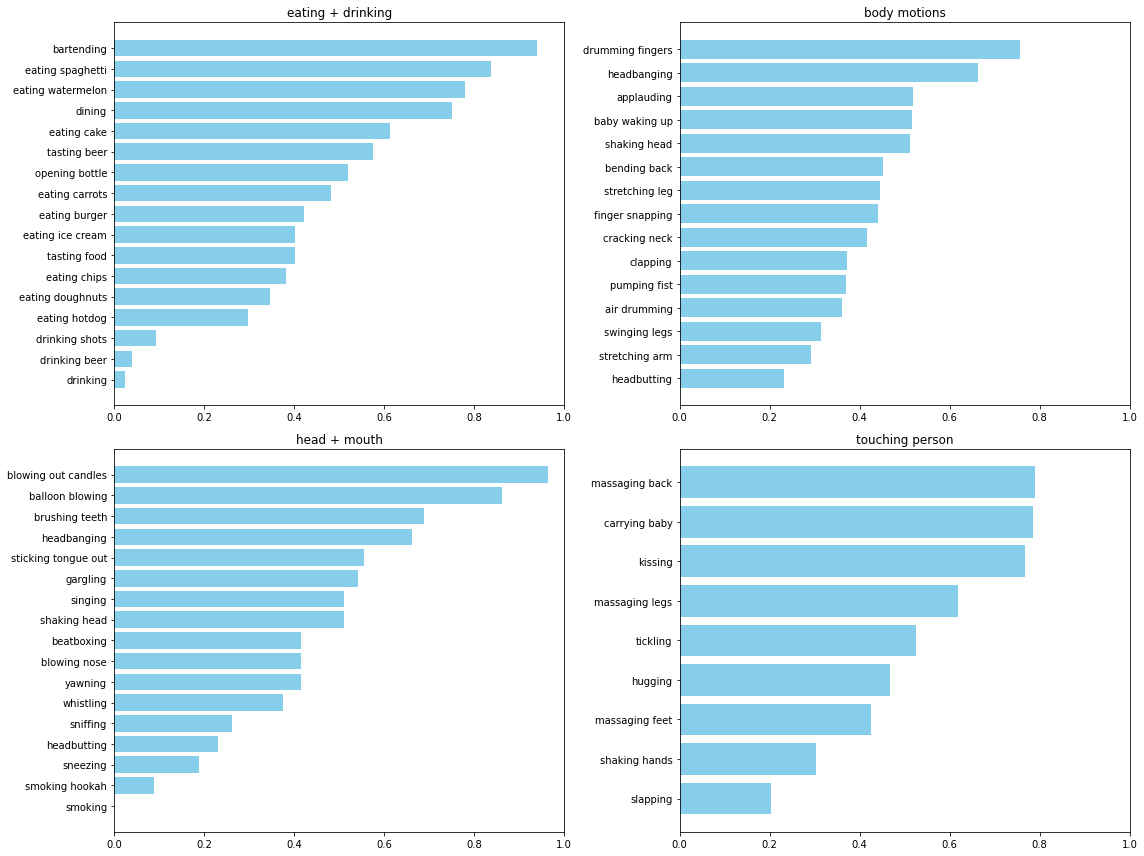} 
	\caption{\textbf{Subclasses of parent classes most affected by occlusion on \textsc{OccludeNet}}. The histogram (blue) highlights the relative accuracy drops for each subclass.}
	\label{fig:top4}
\end{figure}

\begin{figure}[t]
	\centering
	\includegraphics[width = \linewidth]{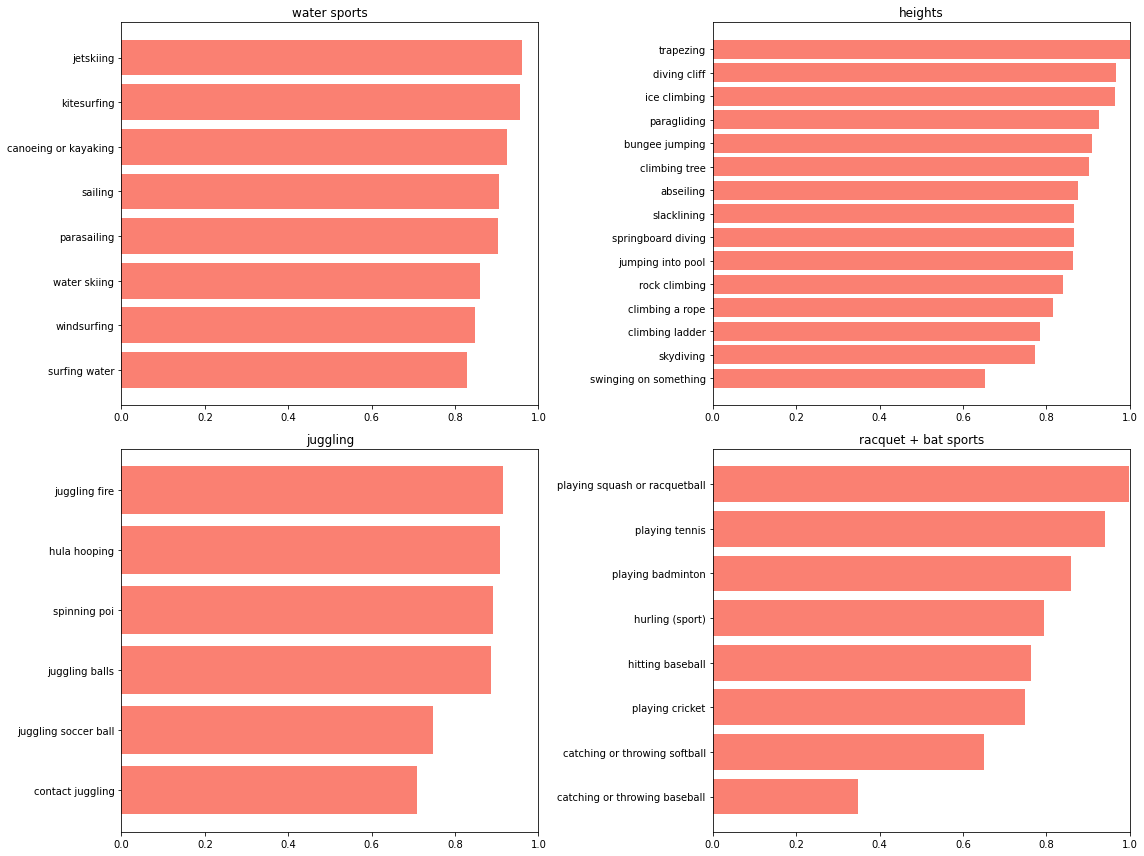} 
	\caption{\textbf{Subclasses of parent classes least affected by occlusion on \textsc{OccludeNet}}. The histogram (red) highlights the relative accuracy changes for each subclass.}
	\label{fig:dtop4}
\end{figure}

\subsubsection{Class Correlation Profiling}

\cref{fig:top4,fig:dtop4} analyze per-class accuracy drops. Actions with low scene relevance or partial body visibility (\textit{e.g.}, eating \& drinking, head movements) exhibit the largest declines under occlusion. In contrast, background-dependent classes (\textit{e.g.}, water sports, racquet \& bat sports) leverage context cues and are less impacted.

\subsection{Diversity and Statistics of Occluded Dataset}
\label{sec:3.2}

\subsubsection{\textsc{OccludeNet-D}}

We randomly sample one-third of \textsc{Kinetics-400} clips and generate three occluded variants per clip, with occluder scales at 0.25, 0.50, and 0.75 of the actor's bounding box. Occluders, backpacks (153), handbags (118), suitcases (888), dogs (1,629), are drawn from COCO~\cite{eccv/LinMBHPRDZ14}, yielding 233,769 clips across 400 classes.

\subsubsection{\textsc{OccludeNet-S}}

We curate 256 clips from \textsc{UCF-Crime}~\cite{cvpr/SultaniCS18} featuring static occlusions, trimmed to match \textsc{Kinetics-400}~\cite{corr/KayCSZHVVGBNSZ17} durations. Classes include abuse (34), burglary (54), fighting (82), robbery (54), and stealing (32). Audio is preserved.

\subsubsection{\textsc{OccludeNet-I}}

Seven action classes, affix poster, armed, basketball dribble, basketball shot, bike ride, indoor run, walk, are recorded with masked participants of varied body shapes and attire. Occlusion modes, levels, and durations differ per clip. Splits: train 80\%, val 5\%, test 15\%.

\subsubsection{\textsc{OccludeNet-M}}

Twelve classes, armed, basketball dribble, basketball shot, bike ride, distribute leaflets, fall, indoor run, outdoor run, steal, throw things, walk, wander, are captured from three views, each with unique occlusions. Splits: train 80\%, val 5\%, test 15\%.

\subsection{Dataset Characteristics Analysis}
\label{sec:3.3}

\subsubsection{Preliminary Analysis}

\textsc{OccludeNet} covers object, scene, and view-variation occlusions with levels from light to heavy. It captures dynamic occlusion patterns, abrupt onset, gradual intensification, and intermittent, creating a challenging evaluation environment.

\begin{figure*}[t]
	\centering
	\begin{tabular}{ccc}
	\includegraphics[width = 0.3\linewidth]{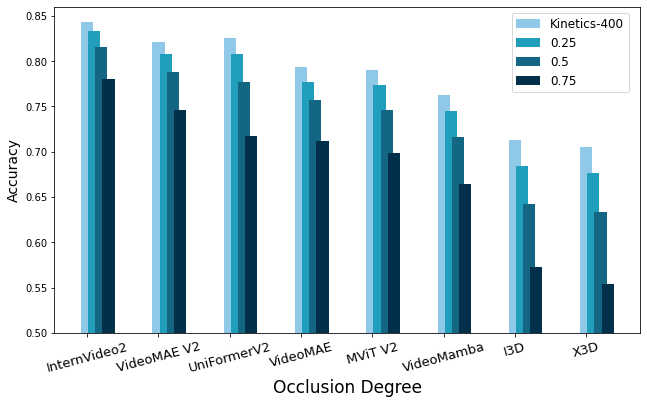} & 
	\includegraphics[width = 0.3\linewidth]{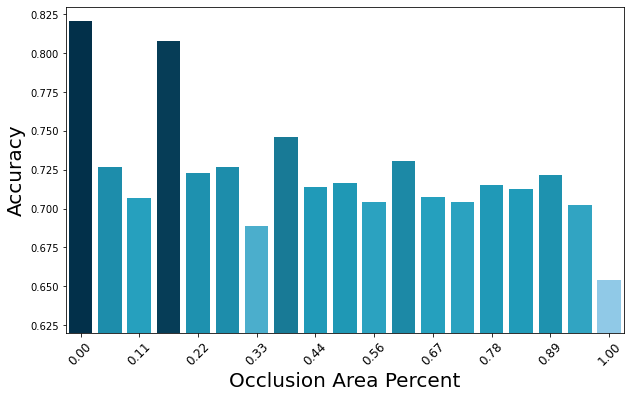} & 
	\includegraphics[width = 0.3\linewidth]{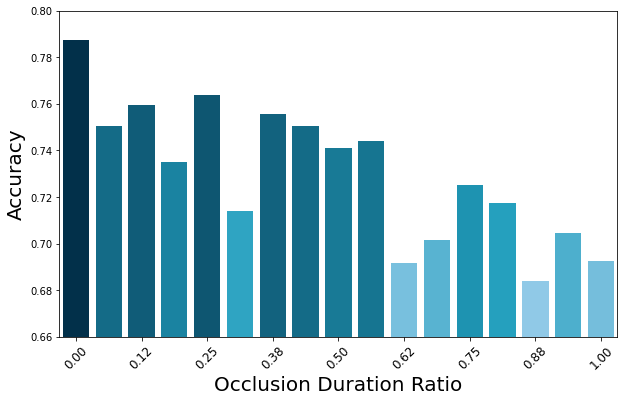} \\
	\small{(a) Occlusion Degree} & \small{(b) Occlusion Area Ratio} & \small{(c) Occlusion Duration Ratio} 
	\end{tabular}
	\caption{\textbf{Impact of individual occlusion factors on recognition accuracy in \textsc{OccludeNet}}. (a) Occlusion degree: ratio of the occluder's size to the actor bounding box; (b) occlusion area ratio: percentage of the actor bounding box that is occluded; (c) occlusion duration ratio: fraction of frames in which the actor is occluded.}
	\label{fig:factors}
\end{figure*}

\subsubsection{Impact of Individual Occlusion Factors}

\cref{fig:factors} defines three metrics:	
\textit{a) Occlusion Degree:} Occluder size relative to actor's bounding box.
\textit{b) Area Ratio:} Percentage of actor bounding box occluded.
\textit{c) Duration Ratio:} Fraction of clip time actor is occluded.






\begin{figure*}[t]
	\centering
	\begin{tabular}{cc}
	\includegraphics[width = 0.5\linewidth]{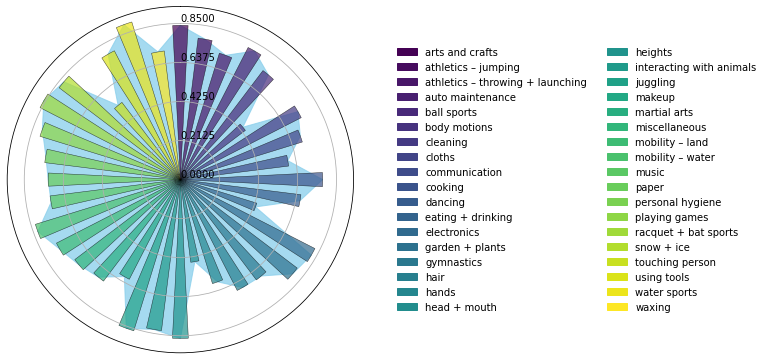} & \includegraphics[width = 0.4\linewidth]{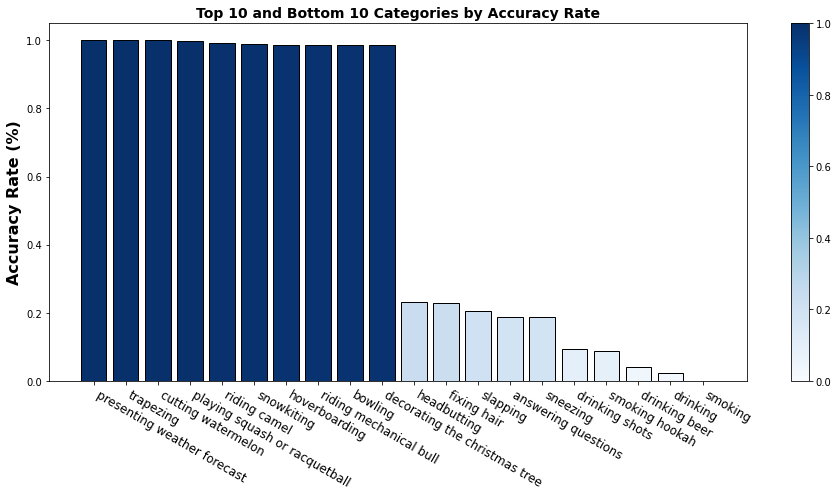} \\
	\small{(a) Parent Classes} & \small{(b) Top 10 Easiest and Hardest Classes}
	\end{tabular}
	\caption{\textbf{Class correlation analysis on \textsc{OccludeNet}}. (a) Overall accuracy for each parent class. (b) Top 10 classes with the highest and lowest accuracy, averaged across all models.}
	\label{fig:interclass}
\end{figure*}

\subsubsection{Class Correlation Profiling}

\cref{fig:interclass} shows that classes with low background relevance or limited visibility (\textit{e.g.}, upper-body actions) experience the largest accuracy drops, underscoring the need for occlusion-aware modeling.

\begin{figure*}[t]
	\centering
	\includegraphics[width = \linewidth]{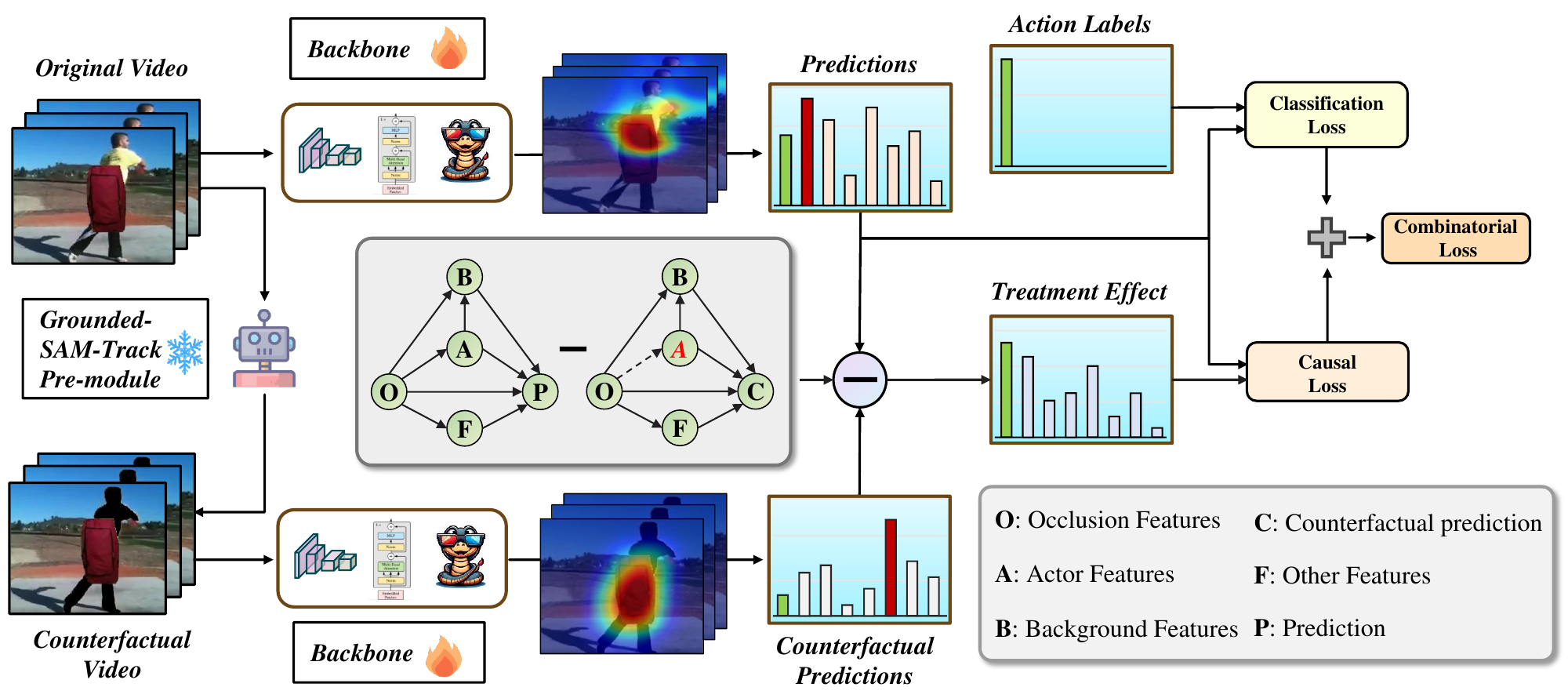}
	\caption{\textbf{Pipeline of CAR}. A preprocessing module generates counterfactual samples that, alongside the original inputs, pass through a shared backbone to produce separate predictions. The discrepancy between these predictions provides a supervised signal via a custom loss, reducing bias from incorrect feature attributions.}
	\label{fig:car}
\end{figure*}

\section{Causal Action Recognition}

We introduce the Causal Action Recognition (CAR) framework, shown in \cref{fig:car}, which models interactions among actors, occlusions, and backgrounds via diverse contextual elements to improve action recognition. This section details CAR's components: occlusion scene modeling (see \cref{sec:4.1}), intervention strategies (see \cref{sec:4.2}), and counterfactual reasoning (see \cref{sec:4.3}).

\subsection{Modeling Occluded Scenes}
\label{sec:4.1}

We represent occluded scenes with a structural causal model (SCM) as a directed acyclic graph, where nodes are variables and edges are causal links. The prediction $P$ is given by
\begin{equation}
	P = f(A, B, F, O, \epsilon_{P}),
\end{equation}
where $f$: mapping function (\textit{e.g.}, neural network modules), $A$: actor features extracted from the visible body parts, $B$: background features capturing scene context, $F$: additional contextual features (\textit{e.g.}, motion cues), $O$: occlusion features encoding the occluder's appearance and location, $\epsilon_{P}$: independent noise term. We further define
\begin{equation}
	B = g(O, \epsilon_{B}), \quad
	F = h(O, \epsilon_{F}), \quad
	A = k(O, \epsilon_{A}),
\end{equation}
so that each feature is influenced by occlusion $O$ and its respective noise. This SCM captures how occlusion affects $P$, forming the basis of CAR's robustness to occlusions.

\subsection{Causal Intervention Strategies}
\label{sec:4.2}

To isolate the causal effect of $A$ on $P$, we apply Pearl's do-calculus~\cite{pearl2009causality}, treating $O$ as a confounder. We compute the interventional prediction	
\begin{equation}
	P\left(\mathrm{do}(A=a)\right),
\end{equation}
which reflects $P$ when $A$ is set to $a$, independent of $O$. We then use back-door adjustment to block $O$'s influence:	
\begin{equation}
	P\left(P\mid\mathrm{do}(A)\right) = \sum_{o} P(P \mid A, o) P(o).
\end{equation}
Implementation relies on precise actor segmentation and tracking via Grounded-SAM~\cite{liu2023grounding, ren2024grounded}, Segment-and-Track-Anything~\cite{corr/abs-2305-06558, iccv/KirillovMRMRGXW23}, and EfficientVit-SAM~\cite{cai2022efficientvit}, enabling controlled manipulation of $A$.


\begin{table*}[t]
	\centering
	\setlength{\tabcolsep}{10pt}
	\caption{\textbf{Top-1 accuracy (\%) of various models on occluded datasets}. D-25, D-50, and D-75 denote \textsc{OccludeNet-D} occlusion levels (25\%, 50\%, 75\%). K-O, O-S, O-I, and O-M refer to \textsc{K-400-O}, \textsc{OccludeNet-S}, \textsc{OccludeNet-I}, and \textsc{OccludeNet-M}, respectively. ``K-400'' indicates performance on the original \textsc{Kinetics-400}.}
	\begin{tabular}{lcc|cccccccc}
	\toprule[1.1pt]
	Method & Venue & Backbone & K-O & D-25 & D-50 & D-75 & O-S & O-I & O-M & K-400 \\
	\midrule
	I3D~\cite{cvpr/CarreiraZ17}	& CVPR'17	 & ResNet-50	& 33.6 & 68.48 & 64.26 & 57.26 & 15.09 & 10.19 & 9.86	& 71.28 \\
	X3D~\cite{feichtenhofer2020x3d}	& CVPR'20	 & ResNet-50	& 34.3 & 67.60 & 63.34 & 55.43 & 16.98 & 19.44 & 2.90	& 70.47 \\
	MViTv2~\cite{cvpr/LiW0MXMF22}	 & CVPR'22	 & Transformer	& 56.7 & 77.35 & 74.64 & 69.83 & 26.42 & 24.07 & 1.16	& 79.01 \\
	VideoMAE~\cite{nips/TongS0022}	& NeurIPS'22& Transformer	& 58.1 & 77.67 & 75.72 & 71.14 & 32.08 & 3.70	& 4.06	& 79.34 \\
	UniFormerv2~\cite{iccv/LiWH0WW023}	& ICCV'23	 & Transformer	&	-	 & 80.82 & 77.68 & 71.79 & 20.75 & 7.41	& 11.30 & 82.60 \\
	VideoMAEv2~\cite{cvpr/WangHZTHWWQ23}	& CVPR'23	 & Transformer	&	-	 & 80.81 & 78.85 & 74.66 & 7.55	& 2.78	& 17.39 & 82.15 \\
	VideoMamba~\cite{li2024videomamba}	& ECCV'24	 & Mamba	&	-	 & 74.54 & 71.68 & 66.43 & 22.96 & 22.53 & 10.55 & 76.28 \\
	InternVideo2~\cite{wang2024internvideo2}& arXiv'24 & Transformer	&	-	 & 83.33 & 81.61 & 78.04 & 20.12 & 16.19 & 3.61	& 84.34 \\
	\bottomrule[1.1pt]
	\end{tabular}
	\label{table2}
\end{table*}

\subsection{Counterfactual Reasoning and Learning}
\label{sec:4.3}

Counterfactual reasoning compares original and intervened predictions to quantify $A$'s causal effect:	
\begin{equation} 
	P_{\mathrm{effect}} = \mathbb{E}_{A \sim \tilde{A}} \left[P(A = \bm{A}, O = \bm{O}) - P\left(\mathrm{do}(A = \bm{a}), O = \bm{O})\right)\right],
\end{equation}
where $\tilde{A}$ is the manipulated feature distribution, $\bm{A}$ is the observed actor state, $\bm{a}$ is the counterfactual state. Denote logits for original and counterfactual predictions by $p_i$ and $c_i$. The softmax probabilities are	
\begin{equation} 
	P = \left(\frac{\exp(p_1)}{\sum_{j=1}^{n}\exp(p_j)}, \frac{\exp(p_2)}{\sum_{j=1}^{n}\exp(p_j)}, \dots, \frac{\exp(p_n)}{\sum_{j=1}^{n}\exp(p_j)} \right),
\end{equation} 
ensuring that $P$ sums to 1 across all classes. The corrected prediction $Y$ is derived from the difference between the logits:
\begin{equation}
	Y = \left(\frac{\exp(p_i - c_i)}{\sum_{j=1}^{n}\exp(p_j - c_j)} \right)_{i = 1}^n.
\end{equation} 
We combine cross-entropy and causal (Kullback-Leibler divergence) losses:	
\begin{equation} 
	\mathcal{L} = -\sum_{i=1}^{n}P_i\log\hat{P}_i + \alpha \sum_{i=1}^{n}P_i\log\frac{P_i}{Y_i},
\end{equation}
where $P_i$ and $Y_i$ are defined as above (original and counterfactual softmax probabilities); $\hat{P}_i$ is the one-hot ground-truth label for class $i$; $\alpha$ is a balancing hyperparameter that weights the causal (KL) term relative to the cross-entropy term. This counterfactual supervision during fine-tuning enhances occlusion robustness without added inference cost.

\section{Experimental Results}

\subsection{Benchmarking and Analysis}
\label{sec:5.1}
\label{sec:syn}

\paragraph{Experimental Setup}

We evaluate eight state-of-the-art models on \textsc{Kinetics-400}~\cite{corr/KayCSZHVVGBNSZ17}, \textsc{K-400-O}~\cite{nips/GroverVR23}, and our \textsc{OccludeNet} test set. The models are InternVideo2~\cite{wang2024internvideo2}, VideoMamba~\cite{li2024videomamba}, UniFormerv2~\cite{iccv/LiWH0WW023}, VideoMAEv2~\cite{cvpr/WangHZTHWWQ23}, VideoMAE~\cite{nips/TongS0022}, MViTv2~\cite{cvpr/LiW0MXMF22}, X3D~\cite{feichtenhofer2020x3d}, and I3D~\cite{cvpr/CarreiraZ17}. We use the mmaction2 framework~\cite{2020mmaction2} and pretrained weights for all models except VideoMamba and InternVideo2. \cref{table2} summarizes Top-1 and Top-5 accuracies across datasets.

\paragraph{Influencing Factors Analysis}

Across all models, accuracy steadily declines as occlusion degree increases (see \cref{fig:factors}(a)). CNN-based models (I3D, X3D) suffer the steepest drops. Similarly, \cref{fig:factors}(b) shows that larger occlusion area ratios, \textit{i.e.}, greater occluder coverage of the actor's bounding box, lead to further accuracy loss. Finally, longer occlusion durations (see \cref{fig:factors}(c)) reduce recognition performance, since fewer visible frames remain.

\paragraph{Parent-Class Impact}

We group actions by parent classes (\textit{e.g.}, ``Music'' includes playing drums, trombone, violin) as in~\cite{corr/KayCSZHVVGBNSZ17}. \cref{fig:interclass}(a) shows that occlusion most severely affects body-motion, eating \& drinking, head \& mouth, and ``touching persons'' classes. These actions rely heavily on body parts occupying large frame areas with minimal background cues, making them vulnerable when occluded.

\paragraph{Class-Specific Characteristics}

\cref{fig:interclass}(b) details per-class accuracy changes on \textsc{OccludeNet-D}. Diet-related actions (eating burgers, watermelon), head-focused actions (brushing teeth, curling hair), hand-centric actions (nail clipping, clapping, sign language), and instrumental actions (playing flute, saxophone) show large drops, as models depend on actor features. In contrast, background-dependent actions (swimming, surfing, rock climbing) are less impacted, leveraging contextual cues.

\begin{table}[t]
	\centering
	\setlength{\tabcolsep}{9pt}
	\caption{\textbf{Ablation study on occlusion degree in \textsc{OccludeNet}}. Unoccluded, Slight, Moderate, and Heavy denote increasing occlusion levels.}
	\begin{tabular}{l|cccc}
	\toprule[1.1pt]
	Metric & Unoccluded & Slight & Moderate & Heavy \\
	\midrule
	Top‐1 (\%) & 77.30 & 75.32 & 72.31 & 66.65 \\
	Top‐5 (\%) & 93.00 & 91.89 & 89.87 & 85.69 \\
	\bottomrule[1.1pt]
	\end{tabular}
	\label{table:occluder}
\end{table}

\paragraph{Generalization of Synthetic Data}

We perform an ablation using UniFormerv2~\cite{iccv/LiWH0WW023} (see \cref{table:occluder}). Although synthetic occlusions reduce model accuracy, human observers still recognize the actions, underscoring the challenge. 

\begin{figure}[t]
	\centering
	\includegraphics[width = \linewidth]{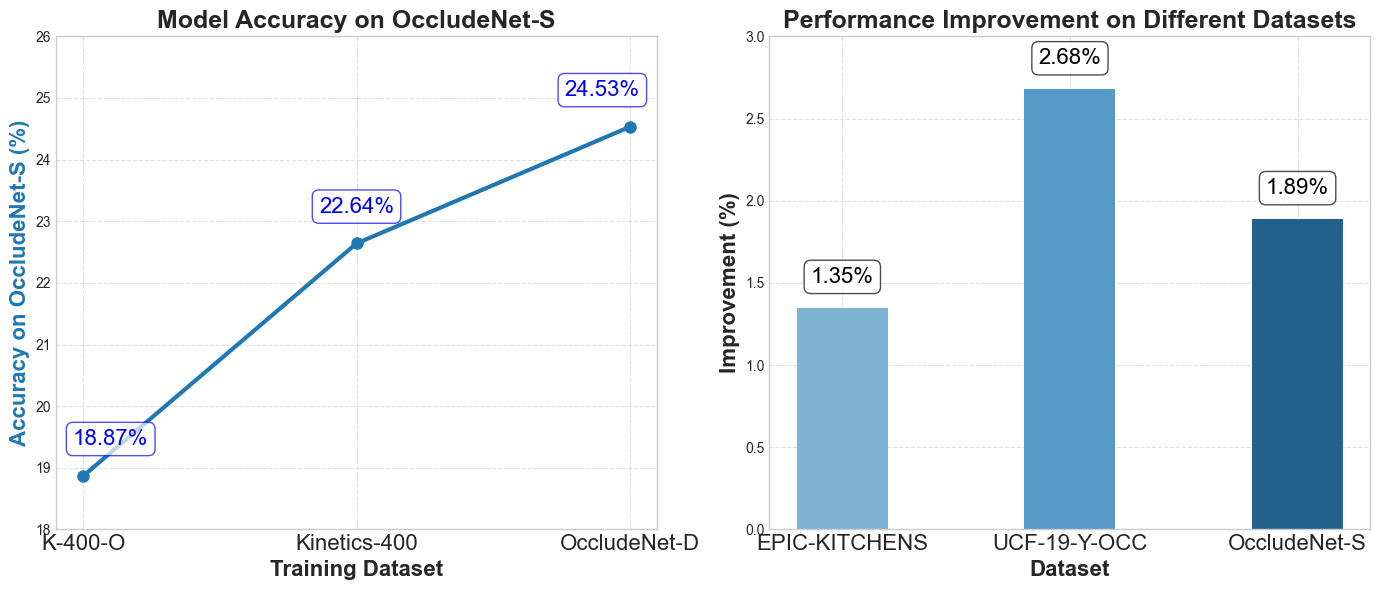}
	\caption{\textbf{Generalization evaluation of \textsc{OccludeNet-D}}. \textbf{Left:} A model trained on \textsc{OccludeNet-D} achieves peak accuracy (24.53\%) on real-world occluded data. \textbf{Right:} Incorporating \textsc{OccludeNet-D} in training yields performance gains across multiple datasets.}
	\label{fig:general}
\end{figure}

Training on \textsc{OccludeNet-D} and testing on real occluded data (see \cref{fig:general}) shows that synthetic-trained models achieve higher accuracy across datasets, demonstrating improved robustness and generalization.

\begin{figure}[t]
	\centering
	\includegraphics[width = \linewidth]{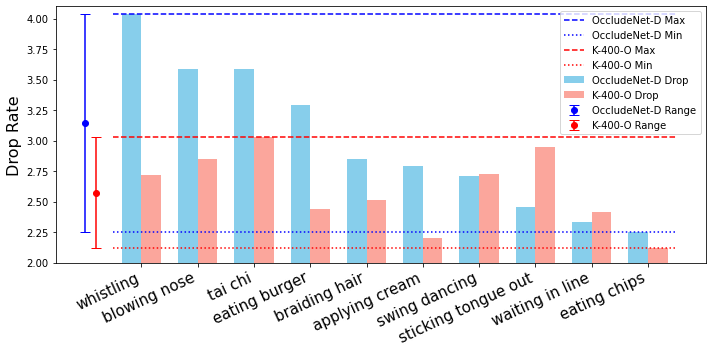}
	\caption{\textbf{Impact of occlusion strategy on recognition accuracy}. For VideoMAE~\cite{nips/TongS0022}, MViTv2~\cite{cvpr/LiW0MXMF22}, I3D~\cite{cvpr/CarreiraZ17}, and X3D~\cite{feichtenhofer2020x3d}, \textsc{OccludeNet-D} induces Top-1 accuracy drops at occlusion levels of 0.25, 0.50, and 0.75, compared to \textsc{K-400-O}. The ten classes with the largest drops on \textsc{OccludeNet-D} were selected at random.}
	\label{fig:9}
\end{figure}

\paragraph{Dataset Sustainability Discussion}

\cref{fig:9} compares inter-class accuracy variance on \textsc{K-400-O} versus \textsc{OccludeNet-D}. Heavy, blind occlusions in \textsc{K-400-O} render recognition nearly impossible, even for humans. By contrast, \textsc{OccludeNet-D} retains sufficient visual cues for human recognition, supporting sustainable accuracy improvements and robustness under occlusion.

\subsection{Causal Action Recognition}
\label{sec:5.2}

To assess causal modeling, we compare CAR against background-modeling baselines (StillMix~\cite{li2023mitigating}, FAME~\cite{cvpr/DingLYQXCWX22}). We fine-tune UniFormerv2-B/16~\cite{iccv/LiWH0WW023} (CLIP-400M+K710 pretrain) for 5 epochs on \textsc{OccludeNet}. For \textsc{OccludeNet-S}, we fine-tune X3D-S~\cite{feichtenhofer2020x3d}, UniFormerv2-B, and VideoMamba-M~\cite{li2024videomamba} for 5 and 10 epochs using both baseline and CAR losses.

\begin{table*}[t]
	\centering
	\setlength{\tabcolsep}{6pt}
	\caption{\textbf{Top‐1 and Top‐5 accuracy (\%) of advanced methods on occluded and standard datasets}. D-25, D-50, D-75 denote \textsc{OccludeNet-D} occlusion levels (25\%, 50\%, 75\%). \textsc{EPIC} refers to the preprocessed Epic-Kitchens clips. StillMix~\cite{li2023mitigating} and FAME~\cite{cvpr/DingLYQXCWX22} are not applicable to \textsc{K-400-O} or Epic-Kitchens.}
	\begin{tabular}{lc|cccccc}
	\toprule[1.1pt]
	Method	 & Backbone	 & D-25	 & D-50	 & D-75	 & \textsc{Kinetics-400} & \textsc{K-400-O} & Epic-Kitchens \\
	\midrule
	StillMix~\cite{li2023mitigating}	& Swin Transformer	 & 55.35 / 79.59 & 53.72 / 78.55 & 51.63 / 76.27 & 57.63 / 81.21	 & -	 & -	 \\
	FAME~\cite{cvpr/DingLYQXCWX22}	 & Swin Transformer	 & 55.13 / 79.88 & 54.21 / 78.72 & 51.64 / 76.33 & 57.41 / 80.98	 & -	 & -	 \\
	UniFormerv2~\cite{iccv/LiWH0WW023} & CNN + Transformer	& 88.37 / 97.69 & 87.16 / 97.25 & 85.13 / 96.50 & 88.31 / 97.71	 & 57.75 / 77.47	& 35.81 / 79.05	\\
	\rowcolor{gray!20}
	UniFormerv2 + CAR	& CNN + Transformer	& \textbf{88.63 / 97.77} & \textbf{87.61 / 97.74} & \textbf{85.50 / 96.66} & \textbf{88.98 / 97.99} & \textbf{57.75 / 77.80} & \textbf{37.16 / 81.76} \\
	\bottomrule[1.1pt]
	\end{tabular}
	\label{table3}
\end{table*}

\begin{table}[t]
	\centering
	\setlength{\tabcolsep}{10pt}
	\caption{\textbf{Top-1 accuracy (\%) of baseline \textit{vs.} CAR-enhanced models on \textsc{OccludeNet-S}, \textsc{OccludeNet-I}, and \textsc{OccludeNet-M}}. S-5/-10: trained for 5/10 epochs on \textsc{OccludeNet-S}; O-I/-M: \textsc{OccludeNet-I}/\textsc{OccludeNet-M}. ``$\Delta$'' denotes improvement with CAR.}
	\begin{tabular}{l|cccc}
	\toprule[1.1pt]
	Method	 & S-5	 & S-10	& O-I	 & O-M	 \\
	\midrule
	X3D~\cite{feichtenhofer2020x3d}	 
		 & 26.42 & 20.75 & 52.78 &	6.09 \\
	\rowcolor{gray!20}
	X3D + CAR	& 26.42 & 28.30 & 54.63 &	8.70 \\
	$\Delta$		& +0.00 & +7.55 & +1.85 & +2.61 \\
	\midrule
	UniFormerv2~\cite{iccv/LiWH0WW023}	
		 & 79.25 & 84.91 & 14.81 & 19.13 \\
	\rowcolor{gray!20}
	UniFormerv2 + CAR	& 81.13 & 86.78 & 15.74 & 23.77 \\
	$\Delta$		& +1.88 & +1.87 & +0.93 & +4.64 \\
	\midrule
	VideoMamba~\cite{li2024videomamba}	
		 & 26.25 & 33.64 & 22.53 & 32.46 \\
	\rowcolor{gray!20}
	VideoMamba + CAR	 & 27.04 & 34.75 & 24.92 & 34.49 \\
	$\Delta$		& +0.79 & +1.11 & +2.39 & +2.03 \\
	\bottomrule[1.1pt]
	\end{tabular}
	\label{table4}
\end{table}

\paragraph{Results and Generalization}

\cref{table3,table4} report accuracies on \textsc{OccludeNet}, \textsc{Kinetics-400}, \textsc{K-400-O}, and \textsc{Epic-Kitchens}~\cite{Damen2018EPICKITCHENS}. CAR consistently outperforms baselines, mitigating occlusion-induced degradation.

\begin{figure}[t]
	\centering
	\includegraphics[width = \linewidth]{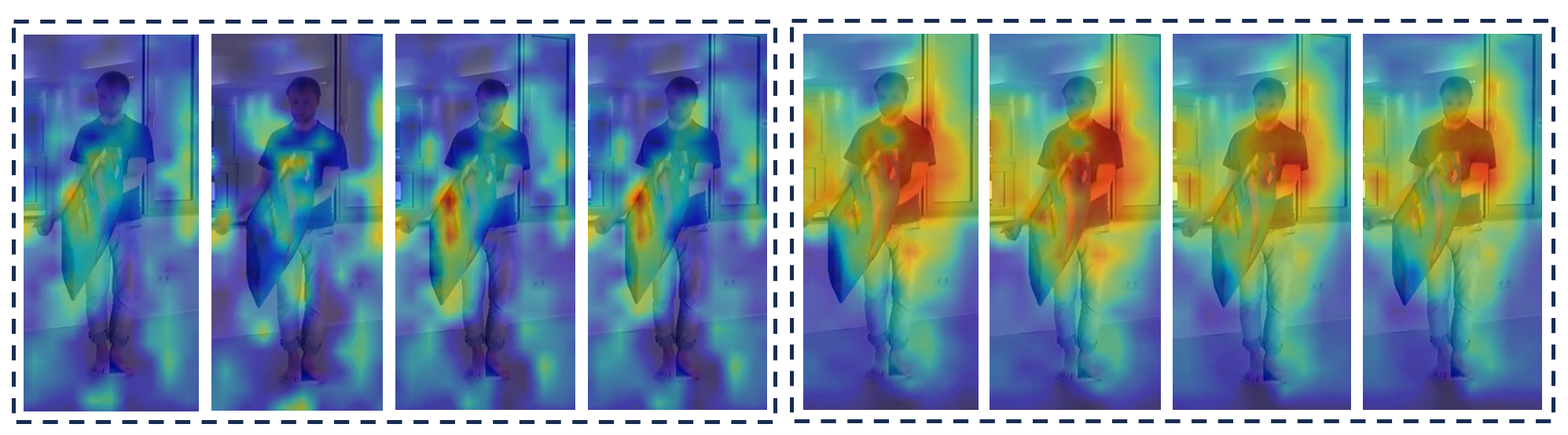} 
	\caption{\textbf{Grad-CAM class activation map comparison~\cite{Selvaraju_2019}}. Baseline (left) versus CAR (right) illustrates that CAR shifts attention from occluders to the actor's action regions, improving focus on relevant features and boosting recognition accuracy.}
	\label{fig:8}
\end{figure}

Grad-CAM visualizations (see \cref{fig:8}) confirm that CAR redirects attention from occluders and backgrounds to the actor's key regions, even on unobstructed \textsc{Kinetics-400} videos.

\begin{table}[t]
	\centering
	\setlength{\tabcolsep}{10pt}
	\caption{\textbf{Ablation on $\alpha$ for CAR’s Top‐1/Top‐5 accuracy (\%) on \textsc{OccludeNet-D}}.}
	\begin{tabular}{c|ccc}
	\toprule[1.1pt]
	$\alpha$ & D-25		& D-50		& D-75		\\
	\midrule
	0	& 88.37 / 97.69 & 87.16 / 97.25 & 85.13 / 96.50 \\
	0.5	& 88.18 / 97.72 & 87.36 / 97.46 & 85.23 / 96.56 \\
	\rowcolor{gray!20}
	1.0	& \textbf{88.63 / 97.77} & \textbf{87.61 / 97.74} & \textbf{85.50 / 96.66} \\
	2.0	& 88.24 / 97.75 & 73.02 / 93.65 & 85.19 / 96.57 \\
	\bottomrule[1.1pt]
	\end{tabular}
	\label{table5}
\end{table}

\paragraph{Causal Inference Ablation.}
We vary the loss balance hyperparameter $\alpha$ (see \cref{table5}) to evaluate its effect on robustness. Results indicate that an optimal $\alpha$ range significantly enhances performance under occlusion.

\begin{figure}[t]
	\centering
	\includegraphics[width = \linewidth]{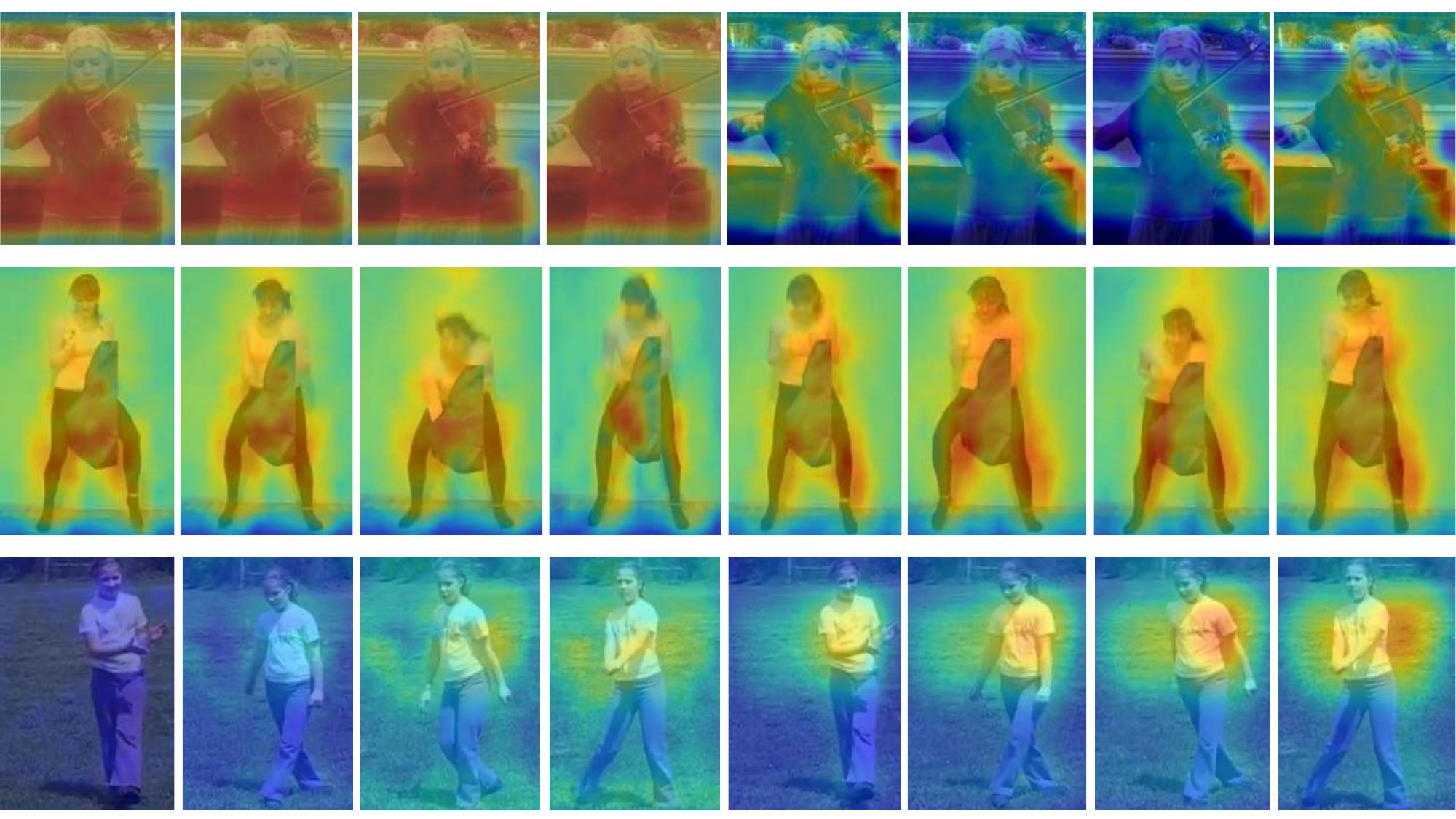} 
	\caption{\textbf{Comparison of class activation maps}. Grad-CAM~\cite{Selvaraju_2019} visualizations for the baseline (left) and CAR (right) models. CAR redirects attention from occluders to the actor's action regions and, even on non-occluded \textsc{Kinetics-400} samples, produces more focused attention on the actor's movements.}
	\label{fig:grad}
\end{figure}

\paragraph{Grad-CAM Visualization}

\cref{fig:grad} shows Grad-CAM attention maps~\cite{Selvaraju_2019}. On \textsc{OccludeNet} dataset, CAR focuses strongly on unoccluded actor regions and key action cues, whereas the baseline model often attends to occluders. Even on the unconstrained \textsc{Kinetics-400} videos, CAR produces more concentrated attention on the actor's movements. These results confirm that CAR not only bolsters occlusion robustness but also refines the model's attention distribution.

\paragraph{Comparison with Actor-Mask Training.}

We have also tried to use actor masks directly for auxiliary training, but due to data diversity, it often lacks continuity. Retaining only the body may fail to convey the action class, while the background and occluders, no longer featuring the actor's action, tend to be more continuous. This reflects the necessity of our causal modeling. The specific experimental data are in tab~\ref{tab:actor}.

\begin{table}[t]
	\centering
	\setlength{\tabcolsep}{9pt}
	\caption{\textbf{Comparison with actor‐mask training}. CAR outperforms direct actor‐mask supervision across occlusion levels.}
	\begin{tabular}{l|ccc}
	\toprule[1.1pt]
	Method						& D-25			 & D-50			 & D-75			 \\
	\midrule
	Actor mask					& 88.21 / 97.65	& 87.10 / 97.21	& 85.13 / 96.46	\\
	\rowcolor{gray!20}
	CAR (ours)		 & \textbf{88.63 / 97.77} & \textbf{87.61 / 97.74} & \textbf{85.50 / 96.66} \\
	\bottomrule[1.1pt]
	\end{tabular}
	\label{tab:actor}
\end{table}

\section{Conclusion}
We introduce \textsc{OccludeNet}, a large-scale video dataset simulating dynamic tracking, static scene, and multi-view interactive occlusions. Our analysis reveals that occlusion strategies impact action classes unevenly, with the greatest accuracy drops for actions featuring low scene relevance and partial body visibility. To mitigate this, we propose the CAR framework, which models occluded scenes via a structural causal model and leverages back-door adjustment and counterfactual reasoning to refocus on unoccluded actor features, significantly improving robustness.

\textbf{Applications:}	
\textit{1) Security and surveillance:} Surveillance footage often suffers from limited viewpoints and intentional obstructions. \textsc{OccludeNet} provides realistic occlusion scenarios to train and evaluate models for more reliable monitoring.	
\textit{2) Virtual scenes:} In short-form and live videos, virtual occlusions such as subtitles, stickers, and tracking effects abound. Our dataset can simulate these artifacts, guiding methods to handle them effectively.	
\textit{3) Occlusion localization:} By exposing models to varied occlusions, \textsc{4) OccludeNet} enables future work on pinpointing occluded regions within frames, advancing spatial understanding.	
\textit{5) Multi-modal analysis:} All clips include audio, supporting integrated audio-visual learning and reflecting the trend toward multi-modal video understanding.

\balance


\bibliographystyle{IEEEtran}
\bibliography{main}

\end{document}